\documentclass[10pt,twocolumn,letterpaper]{article}

\usepackage{iccv}
\usepackage{times}
\usepackage{epsfig}
\usepackage{graphicx}
\usepackage{amsmath,nccmath}
\usepackage{amssymb}

\usepackage{subcaption}
\usepackage{amsbsy}
\usepackage{mathtools}
\usepackage{enumitem}
\usepackage{eucal}
\usepackage{booktabs}
\renewcommand{\restriction}{\mathord{\upharpoonright}}

\usepackage[pagebackref=true,breaklinks=true,colorlinks,bookmarks=false]{hyperref}

\iccvfinalcopy 

\def\iccvPaperID{17} 

\ificcvfinal\fi

\begin{document}

\title{Part-to-whole Registration of Histology and MRI using Shape Elements}

\author{Jonas Pichat$^1$
\and Juan Eugenio Iglesias$^1$
\and Sotiris Nousias$^1$
\and Tarek Yousry$^2$
\and S\'ebastien Ourselin$^{1,3}$ \ \ \ Marc Modat$^1$\\
$^1$Translational Imaging Group, CMIC, University College London, UK\\
$^2$Department of Brain Repair \& Rehabilitation, UCL Institute of Neurology, UK\\
$^3$Wellcome / EPSRC Centre for Interventional and Surgical Sciences, UCL, UK\\
{\tt\small jonas.pichat.13@ucl.ac.uk}
}

\maketitle
\thispagestyle{empty}

\begin{abstract}
Image registration between histology and magnetic resonance imaging (MRI) is a challenging task due to differences in structural content and contrast.  Too thick and wide specimens cannot be processed all at once and must be cut into smaller pieces. This dramatically increases the complexity of the problem, since each piece should be individually and manually pre-aligned. To the best of our knowledge, no automatic method can reliably locate such piece of tissue within its respective whole in the MRI slice, and align it without any prior information. We propose here a novel automatic approach to the joint problem of multimodal registration between histology and MRI, when only a fraction of tissue is available from histology. The approach relies on the representation of images using their level lines so as to reach contrast invariance. Shape elements obtained via the extraction of bitangents are encoded in a projective-invariant manner, which permits the identification of common pieces of curves between two images. We evaluated the approach on human brain histology and compared resulting alignments against manually annotated ground truths. Considering the complexity of the brain folding patterns, preliminary results are promising and suggest the use of characteristic and meaningful shape elements for improved robustness and efficiency.
\end{abstract}


\begin{figure}
\centering
\includegraphics[width=0.47\textwidth]{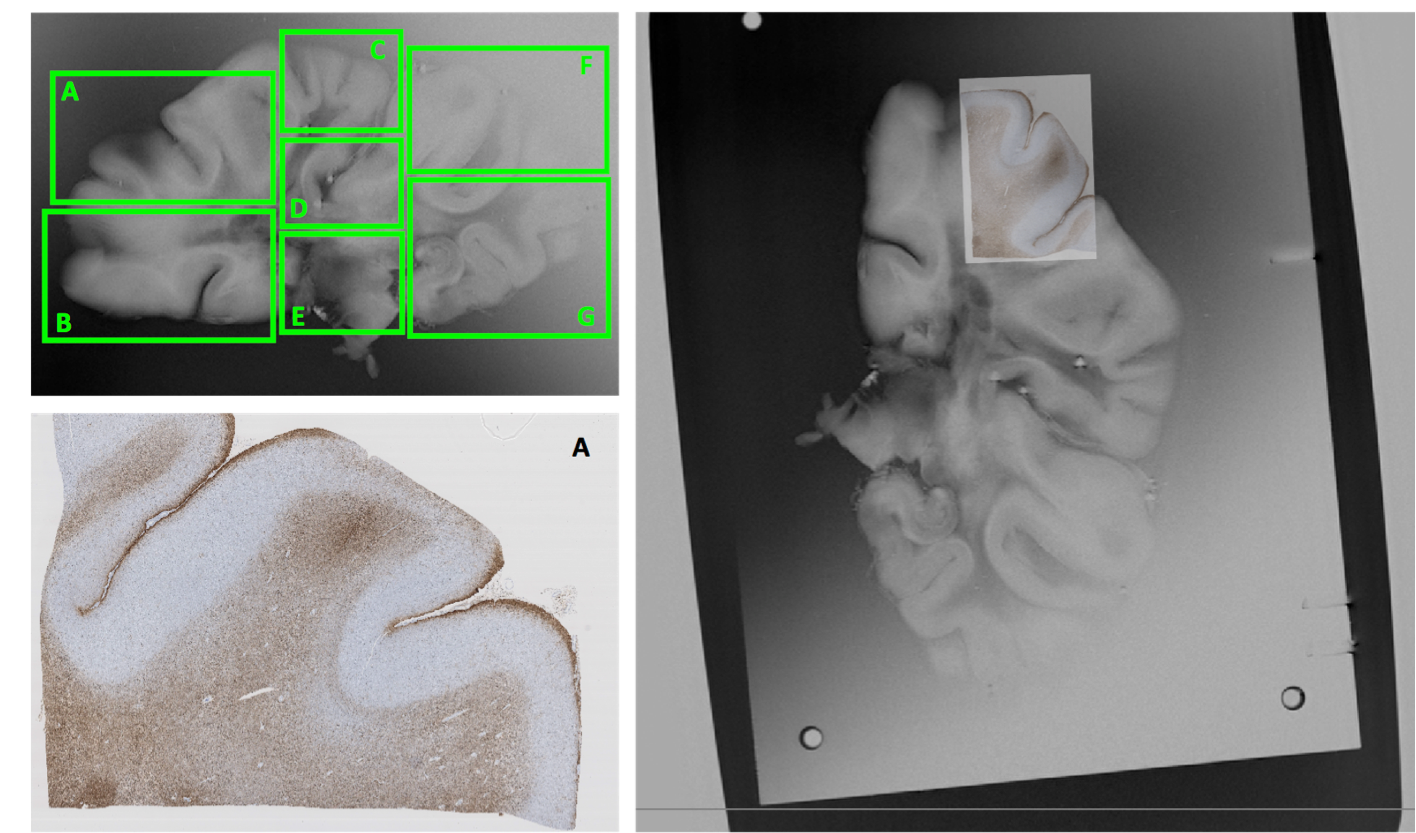}
\caption{Given a histological slice (bottom left) part of a whole specimen, our method aims to automatically spot where it was taken from in the clinical image and align it (right). The result should agree with the areas delineated prior to cutting (top left) to avoid any manual intervention.}
\label{fig:fig1}
\end{figure}

\section{Introduction}


Histology is concerned with the various methods of microscopic examination of a thin tissue section. Cutting through a specimen permits the investigation of its internal topography and the observation of complex differentiated structures through staining. 

\noindent MRI constitutes an invaluable resource for routine, accurate, non-invasive study of biological structures in three dimensions. Relative to histology, MRI avoids irreversible damage and distortions induced by processing, cutting, mounting and staining during the histological preparation. However, resolution-wise, it is outperformed by histology.

\noindent One of the many benefits of combining histology and MRI is to confirm non-invasive measures with baseline information on the actual properties of tissues \cite{annese2012importance} by accessing simultaneously the chemical and cellular information of the former and the rich structural information of the latter.

Such combination relies on image registration and this can be achieved using iconic (intensity-based) \cite{adler2014histology} or geometric (feature-based) \cite{khimchenko2016extending} approaches. Unfortunately, the extraction and manipulation of meaningful information from histology and clinical images is a very complicated task because each modality has, by nature, its own features and there does not always exist a mapping between their constituents: local intensity mappings are non-linear and images exhibit different structures---which is also a reason why intensity-based methods tend to get trapped in local optima.
Hence, classical feature description methods, such as SIFT \cite{lowe1999object}, fail to match features \cite{toews2013feature}. Incidentally, manual extraction of landmarks may remain the safest way to establish correspondences across modalities \cite{gangolli2017quantitative}.

Besides, it is common for histopathology laboratories to receive tissue samples that are: (P1) too wide or (P2) too thick, to be processed as they are. The sample is therefore cut into separate sub-blocks, each of which is processed individually. If no scan of each sub-block is available (unlike in \cite{adler2014histology} for example), one must keep track on which part of the sample each sub-block corresponds to and use that knowledge to initialise the registration of histological slices with the clinical image, or manually align them. As for problem (P2), attempts at using similarity measures have been made to initialise registrations, but those are ambiguous and rely on absolute measures rather than relative ones \cite{yang2012mri}. On that matter, it was shown in \cite{xiao2011determining} that direct comparison of images from different modalities is non-trivial, and fails to reliably determine slice correspondences. To the best of our knowledge, no automatic method to address (P1) (see Fig.~\ref{fig:fig1}) has been proposed in the literature.


\subsection{Related work}

Regarding geometric approaches, one possible strategy to align histology and clinical imaging is to simplify the images into their contours, so as to come down to a monomodal registration problem and use the shape information provided by the external boundaries. In \cite{ali1998registering}, contours from both histology and slices from a rat brain atlas were extracted via thresholding and represented using B-splines. Then, they were described by means of sets of affine invariants constructed from the sequence of area patches bounded by the contour and the line connecting two consecutive inflections. In \cite{trahearn2014fast}, Curvature Scale Space \cite{mokhtarian1986scale} was used for the registration of whole-slide images of histological sections in order to represent shape (the tissue boundary) at various scales. In \cite{davatzikos1996image}, curvature maps at different scales were used to match boundaries of full brain MRI extracted via an active contour algorithm. The main weaknesses of active contours are the number of parameters and the sensitivity to initialisation. 

An alternative to using a single contour was proposed by morphologists, observing that level lines (the boundaries of level sets) provide a complete, contrast-invariant representation of images. Furthermore, level lines fit the boundaries of structures and sub-structures of objects very well. Then, given two images, the problem is to retrieve all the level lines that are common to both images; this is however feasible only if curves have been appropriately simplified (smoothed) \cite{guichard2004contrast} (p.95). Like in \cite{lisani2003theory}, smooth pieces of level lines (the \textit{shape elements} \cite{caselles1996kanisza}) can be encoded to represent shape locally in e.g., an affine-invariant manner \cite{lamdan1988object}. The comparison of the resulting canonical curves then permits to identify portions of level lines common to two images. 

Problem (P1) being multimodal and fractional by nature, it seems natural to formulate a solution that involves contrast- and geometric-invariance, as well as locality.

Here, we present a novel approach to (P1) based on: $(i)$ representing both histology and MRI images using their level lines \cite{lisani2003theory}. This allows to reach contrast invariance and to consider implicitly several structural layers of the images---as opposed to relying solely on the outer boundaries of tissues. From there, characteristic shape elements can be extracted locally along the level lines via their bitangents~(\text{\S}\ref{find_bitangents}). $(ii)$ Representing those elements in a projective invariant manner~(\text{\S}\ref{projective_representation}) as introduced by Rothwell in \cite{rothwell1995object}, so as to be robust to some non-linear deformations that tissues undergo during the histological process. Combining the two procedures permits the partial matching of shape elements regardless of the orientation of the tissue on glass slides. Registration is then obtained as a result of shape recognition~(\text{\S}\ref{reg}).

\subsection{Contributions}
	
\begin{enumerate}[nolistsep]
    \item We address the joint problem of multimodal registration between a fraction of histology and its whole in an MRI slice as a result of shape recognition using portions of level lines. 
    \item We introduce an efficient refinement of bitangents via ellipses.
    \item We extend Rothwell's framework to bitangents crossing the level lines and compare the resulting canonical curves using the Fr\'echet distance. 
\end{enumerate}


\section{Preprocessing}
\label{preproc}
\begin{figure}
\centering
\includegraphics[width=0.47\textwidth]{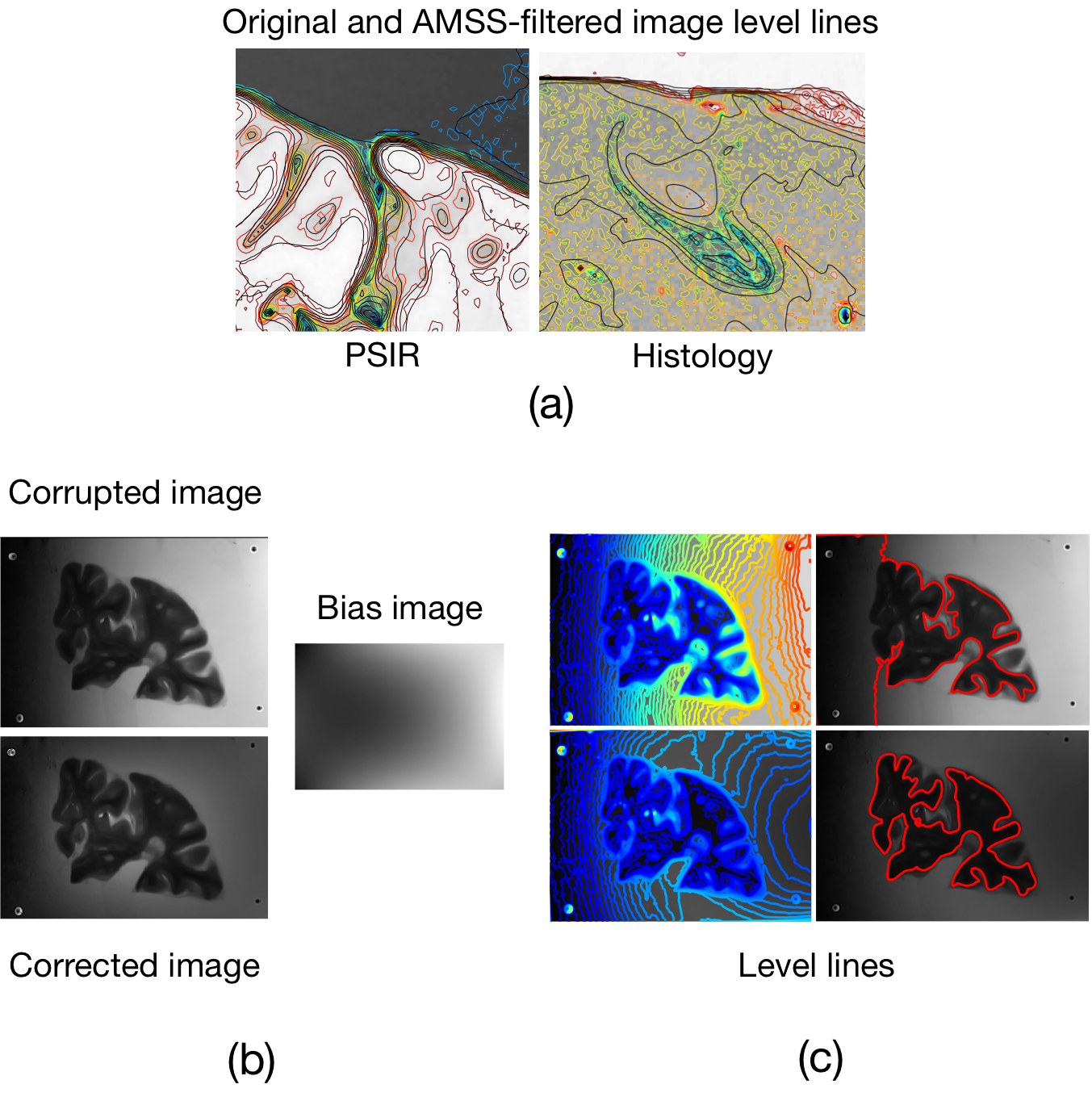}
\caption{Smoothing and bias correction. (a) Level lines (multiple of 16) prior (in colour) and posterior to AMSS filtering (scale 2). (b) Corrupted and bias corrected images of a T2 image of a human brain, along with the estimated bias (Legendre polynomials of order 2). (c) The effect on level lines of corrupted and corrected images is shown.}
\label{fig:fig2}
\end{figure}

We used two standard preprocessing steps: first, smoothing, in order to simplify the image, preserve the shape of the tissue, remove unnecessary details and obtain smooth level lines (Fig.~\ref{fig:fig2}a); then, intensity correction, in order to account for inhomogeneities of the field in MR images (Fig.~\ref{fig:fig2}c) or illumination in histology. 

Smoothing is based on Affine Morphological Scale Space (AMSS) \cite{alvarez1993axioms}. It is governed by the partial differential equation: $\frac{\partial u}{\partial t} = |Du|\textmd{curv}(u)^{1/3}$ where $u$ is the image, $|Du|$ is the gradient of the image, $\textmd{curv}(u)$ is the curvature of the level line and $t$ is a scale parameter. AMSS smoothes homogeneous regions but enhances tissue boundaries. The sequence of updates necessary to its computation follows that presented in ~\cite{mondelli11} (equations of \text{\S}$2.3$).

Image intensities correction relies on surface fitting \cite{dawant1993correction}: the low-frequency bias of an image can be estimated using an adequate basis of smooth and orthogonal polynomial functions. It then comes down to solving the least square problem $A\boldsymbol{\mathsf{c}}=\boldsymbol{\mathsf{b}}$, where $\boldsymbol{\mathsf{b}}\in\mathbb{R}^N$ is the vector of all the pixels values, and $\boldsymbol{\mathsf{c}}$ the coefficients of one linear combination of basis functions. $A \in \mathbb{R}^{N\times (n+1)(m+1)}$ is the matrix of the system: its $k$-th row is the vectorised outer product $\Phi(x_k)\otimes \Phi(y_k)$ with $\Phi(x_k) = [P_0(x_k), P_1(x_k), \dots, P_m(x_k)]^T$ and $\Phi(y_k) = [P_0(y_k), P_1(y_k), \dots, P_n(y_k)]$ for pixel $k\leq N$. $P_i(.)$ denotes a certain 1D polynomial of degree $i$.  Degrees $m$ and $n$ are usually taken small so as not to overfit the image intensities. The left inverse of $A$ (it is full rank) gives the bias image, and correction is straightforward (Fig.~\ref{fig:fig2}b).

\section{Finding bitangents}
\label{find_bitangents}
Characteristic shape elements are extracted by means of bitangents of level lines. Bitangents are identified via the tangent space (\text{\S}\ref{dual_curve}) and each one is refined using two ellipses fitted in the neighbourhood of estimated bitangent points (\text{\S}\ref{ellipse_fit}). Since two ellipses have at most four bitangents (\text{\S}\ref{bitangent_ellipses}), one needs to be singled out which corresponds to the refined bitangent of the level line (\text{\S}\ref{select_bitangent}).

In the following, a \textit{bitangent point} is one of the two points where a bitangent is in contact with the level line. The \textit{length} of a bitangent is defined as the number of inflections of the portion of level line that it covers. As a result, a \textit{short} bitangent refers to a bitangent that covers portions with exactly two inflections and a \textit{long} one, more than two.

\subsection{Dual curve}
\label{dual_curve}

Let $\mathcal{L}$ be a Jordan curve (level lines are plane simple curves, though closedness is not guaranteed for all of them in practice). Duality is defined as the polarity that sends any point to a line and \textit{vice versa}. The image of a point with parameter $t=t_0$ is the line:
\begin{equation}
\label{eq3}
ux(t_0) + vy(t_0) + 1 = 0.
\end{equation}
If the parameter $t$ covers the whole range of definition, the resulting set of straight lines is the envelope of $\mathcal{L}$: the dual $\mathcal{L}^*$ of $\mathcal{L}$ is the set of its tangent lines. A parametrisation of $\mathcal{L}^*$ in homogeneous coordinates can be obtained from~(\ref{eq3}) by differentiation  w.r.t the parameter $t$ and elimination. This yields $u=\frac{-\dot{y}(t)}{\dot{x}(t)y(t) - \dot{y}(t)x(t)}$ and $ v =\frac{\dot{x}(t)}{\dot{x}(t)y(t) - \dot{y}(t)x(t)}$ with $x,y\neq 0$ (dot notation is used for differentiation).

\begin{figure}
\centering
\includegraphics[scale=0.15]{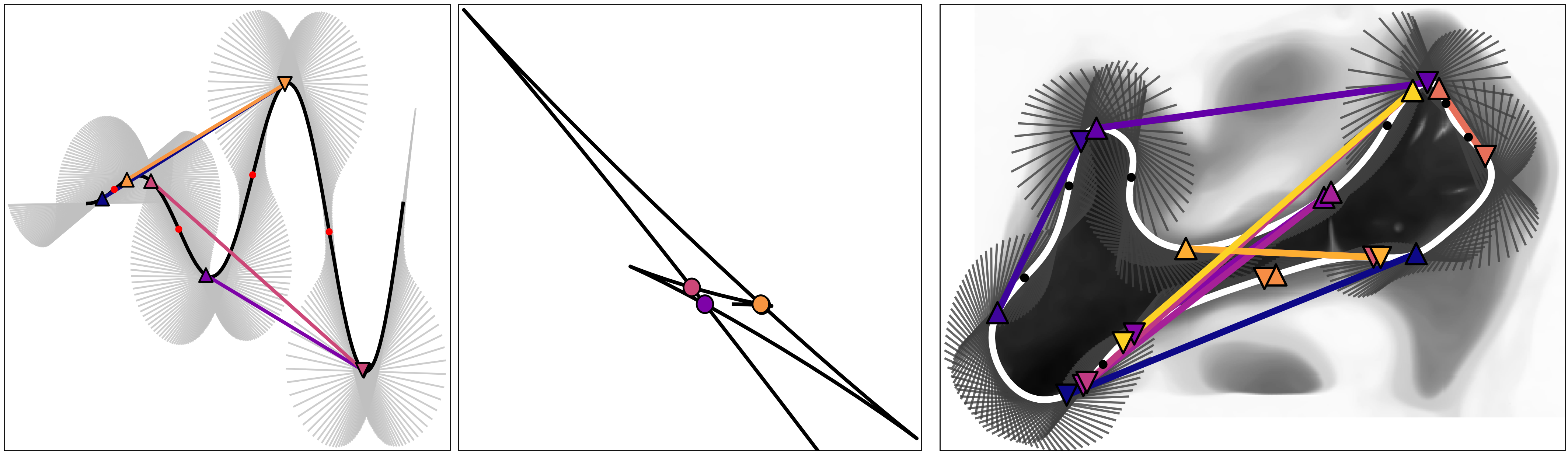}
\caption{Left: the function ${x\mapsto x\textmd{sin}(x)}$ for all $x \in [0,4\pi]$, and the set of its tangents (in grey) are shown. Inflection points are shown with red dots (black dots in the right picture) and bitangents are coloured lines. Middle: the dual curve: its 4 crossing points correspond to the 4 coloured bitangents on the left. Right: bitangents (11 in total) of one level line (in white) from a histological slice (after \text{\S}\ref{dual_curve}).}
\label{fig:fig3}
\end{figure}

Dual curves feature the following properties: an inflection of $\mathcal{L}$ maps onto a cusp of the dual, and two points sharing a common tangent map onto a double point of the dual curve. More generally, a set of $n$ points sharing a common tangent line maps onto a point of multiplicity $n$ of the dual curve. Finding the bitangents of $\mathcal{L}$ is therefore equivalent to finding self-intersections of the polygonal curve $\mathcal{L}^{*}$ (Fig.~\ref{fig:fig3}). To that end, we used the Bentley-Ottmann algorithm \cite{bentley1979algorithms,barton}, which is a line sweep algorithm that reports all intersections among line segments in the plane.

\subsection{Refining bitangents locations}
\label{refine_bitangents}
The refinement of bitangents is preferable: since the slopes of tangents vary substantially in portions of high curvature, the lengths of segments of the dual curve increase on portions where a self-intersection may happen. The evaluation of that double point thus degrades, which directly affects the estimation of bitangents.

\subsubsection{Ellipse fitting}
\label{ellipse_fit}
In order to cope with bitangent errors, we propose to refine their locations by fitting ellipses \cite{fitzgibbon1999direct} around estimated bitangent points. This allows skipping the rotation part prior to the quadratic fitting in \cite{rothwell1995object}. Beforehand, bitangents lying on almost straight edges of the level lines are removed by looking at the residual of a line fit on the portions bounded by the two bitangent points. This is intended to avoid the degenerate case of fitting an ellipse to a nearly straight line.

Let $F$ be a general conic. It is defined as the set of points such that:
\begin{equation}
\label{eq4}
F(\boldsymbol{\mathsf{a}},\mathbf{x}) = \boldsymbol{\mathsf{a}}.\mathbf{x} = ax^2 + by^2 + cxy + dx + ey + f = 0,
\end{equation}

\noindent where $\boldsymbol{\mathsf{a}}=[a\textmd{ }b\textmd{ }c\textmd{ }d\textmd{ }e\textmd{ }f]^T$ and $\mathbf{x}=[x^2\textmd{ }y^2\textmd{ }xy\textmd{ }x\textmd{ }y\textmd{ }1]^T$.

The constrained least square problem we wish to solve here is: $\textmd{min}_{\boldsymbol{\mathsf{a}}} = \boldsymbol{\mathsf{a}}^TS\boldsymbol{\mathsf{a}}$ subject to $\boldsymbol{\mathsf{a}}^TC\boldsymbol{\mathsf{a}}=1$, where ${S=D^TD}$ is the \textit{scatter} matrix, $D$ is the \textit{design} matrix, made of the $N$ points to be fitted and $C$ is the \textit{constraint} matrix which expresses the constraint $4ac - b^2 = 1$ on the conic parameters to make it an ellipse. This translates in a $[6\times 6]$ matrix where $C_{22}=-1$ and  $C_{31}=C_{13}=2$, the rest being zeros. 

This yields the generalised eigenvalue problem (GEP):
\begin{equation}
\label{eq7}
S\boldsymbol{\mathsf{a}} = \lambda C\boldsymbol{\mathsf{a}}.
\end{equation}

The ellipse coefficients, $\boldsymbol{\mathsf{a}}$ are the elements of the eigenvector that corresponds to the only positive eigenvalue. Although the impact of $S$ being nearly singular and $C$ being singular on the stability of the eigenvalues computation is discussed in \cite{halir1998numerically}, we did not encounter any problem in our experiments.

\subsubsection{Bitangents of ellipses}
\label{bitangent_ellipses}
The main goal of this section is to compute the bitangents of two ellipses efficiently. This is achieved by transforming a system of two polynomial equations into a polynomial eigenvalue problem, and for further performance, into a generalised eigenvalue problem.

Let us consider two ellipses, $E_1(\boldsymbol{\mathsf{a_1}},\mathbf{x})$ and $E_2(\boldsymbol{\mathsf{a_2}},\mathbf{x})$ defined by bivariate quadratic polynomials, like in (\ref{eq4}). The tangent line, $T\textmd{:}\textmd{ }y=ux+v$ to say $E_1$, is the line that intersects $E_1$ at exactly one point. By substitution, one gets a degree 2 polynomial in $x$, which has a single root if and only if its discriminant, $\Delta(\boldsymbol{\alpha_1},\boldsymbol{\mathsf{u}})=0$. When considering the tangent to both ellipses, this gives a system of $n=2$ polynomial equations in unknowns $u,v$: 
\begin{equation}
\label{eq8}
(s1)
\begin{cases}
\alpha_{11}u^2 + \alpha_{12}v^2 + \alpha_{13}uv + \alpha_{14}u + \alpha_{15}v + \alpha_{16} = 0\\
\alpha_{21}u^2 + \alpha_{22}v^2 + \alpha_{23}uv + \alpha_{24}u + \alpha_{25}v + \alpha_{26} = 0,
\end{cases}
\end{equation}

\noindent ${\alpha_{i1} = e_i^2 - 4c_if_i\textmd{,  }\alpha_{i2} = b_i^2 - 4a_ic_i\textmd{,  }\alpha_{i3} = 4c_id_i-2b_id_i}$, ${\alpha_{i4} = 2d_ie_i-2b_if_i\textmd{,  }\alpha_{i5} = 2b_id_i-4a_ie_i}\textmd{ and }$ ${\alpha_{i6} = d_i^2-4a_if_i}$, ${i=\{1,2\}}$.

To start with, $u$ is hidden in the coefficient field; $(s1)$ becomes a system of two equations $f_1(u,v)$ and $f_2(u,v)$ in one variable $v$ and coefficients from $\mathbb{R}[u]$ i.e. ${f_1,f_2 \in (\mathbb{R}[u])[v]}$. The degrees of these two equations are $d_1=d_2=2$.

Homogenising $(s1)$ using a new variable $w$ gives $(s2)$, a system of two homogeneous polynomial equations $F_1(v,w)$ and $F_2(v,w)$ in two unknowns $v,w$:
\begin{equation}
\label{eq9}
(s2)
\begin{cases}
\scalebox{0.88}{%
$\alpha_{11}u^2 + \alpha_{12}v^2 + \alpha_{13}uv + \alpha_{14}uw + \alpha_{15}vw + \alpha_{16}w^2 = 0$
}\\
\scalebox{0.88}{%
$\alpha_{21}u^2 + \alpha_{22}v^2 + \alpha_{23}uv + \alpha_{24}uw + \alpha_{25}vw + \alpha_{26}w^2 = 0,$
}
\end{cases}
\end{equation}

The total degree $d=\sum_{i=1}^n(d_i-1)+1$ equals 3. This gives the set $\mathcal{S}$ of $\mbinom{n+d-1}{d}=4$ possible monomials ${\boldsymbol{\mathsf{\omega}}^{\delta}=v^{\delta_2}w^{\delta_3}}$ in variables $v,w$ of total degree $d$ i.e., such that $|\delta|=\sum_{i=2}^3\delta_i=3$: $\mathcal{S}=\{v^3,v^2w, vw^2, w^3\}$. The set $\mathcal{S}$ can be partitioned into two subsets according to a modified Macaulay-based method~\cite{kukelova2013algebraic}:
\begin{equation}
\label{eq10}
\begin{array}{lcl}
\mathcal{S}_1=\{\boldsymbol{\mathsf{\omega}}^{\delta}: |\delta|=3, v^{d_1}|\boldsymbol{\mathsf{\omega}}^{\delta} \},\\
\mathcal{S}_2=\{\boldsymbol{\mathsf{\omega}}^{\delta}: |\delta|=3, w^{d_2}|\boldsymbol{\mathsf{\omega}}^{\delta} \}.
\end{array}
\end{equation}
In other words, $\mathcal{S}_1$ (resp. $\mathcal{S}_2$) is the set of monomials of total degree $3$ that can be divided by $v^2$ (resp. $w^2$). This gives $\mathcal{S}_1=\{v^3, v^2w\}$ and $\mathcal{S}_2=\{vw^2, w^3\}$, from which the extended set of four polynomial equations: ${vF_1 = 0}$, ${wF_1 = 0}$, ${vF_2 = 0}$ and ${wF_2 = 0}$ can be derived.

After dehomogenisation (by setting $w=1$), the extended system can be rewritten as a polynomial eigenvalue problem (PEP):
\begin{equation}
\label{eq11}
\mathsf{C}(u)\boldsymbol{\mathsf{v}}=0,
\end{equation}
where $\boldsymbol{\mathsf{v}}=[v^3\textmd{ }v^2\textmd{ }v\textmd{ }1]^T$ and\\
$\mathsf{C}(u)=
\begin{bsmallmatrix}
    \alpha_{12}  & \alpha_{13}u + \alpha_{15} & \alpha_{11}u^2+\alpha_{14}u + \alpha_{16} & 0 \\
	0 & \alpha_{12}  & \alpha_{13}u + \alpha_{15} & \alpha_{11}u^2+\alpha_{14}u + \alpha_{16} \\
    \alpha_{22}  & \alpha_{23}u + \alpha_{25} & \alpha_{21}u^2+\alpha_{24}u + \alpha_{26} & 0 \\
    0 & \alpha_{22}  & \alpha_{23}u + \alpha_{25} & \alpha_{21}u^2+\alpha_{24}u + \alpha_{26}
\end{bsmallmatrix}$.

Non-trivial solutions to (\ref{eq11}) are the roots of $\textmd{det}(\mathsf{C})$, which gives up to 4 real solutions for $u$. 

For each one of them e.g., $u_1$, the corresponding singular value decomposition has the form: $\mathsf{C}(u_1)=\mathsf{U}\mathsf{S}\mathsf{V}^T$, where the solution vector $[\mathsf{v}_1\textmd{ }\mathsf{v}_2\textmd{ }\mathsf{v}_3\textmd{ }\mathsf{v}_4]^T$ is the column of $\mathsf{V}$ that corresponds to the smallest singular value. The particular solution $v_1$ associated with $u_1$ is e.g. $\frac{\mathsf{v}_3}{\mathsf{v}_4}$, meaning that one bitangent is parametrised by $T_1\textmd{:}\textmd{ }y=u_1x+v_1$.

For the sake of completeness, the PEP (\ref{eq11}) can be further transformed into a GEP by first rewriting it as:
\begin{equation}
\label{eq12}
\scalebox{0.84}{%
$\Bigg(
\underbrace{\begin{bsmallmatrix}
    0  & 0 & \alpha_{11} & 0 \\
	0  & 0 & 0 & \alpha_{11} \\
    0  & 0 & \alpha_{21} & 0 \\
    0  & 0 & 0 & \alpha_{21}
\end{bsmallmatrix}}_\text{$\mathsf{C}_2$} u^2 + 
\underbrace{\begin{bsmallmatrix}
    0  & \alpha_{13} & \alpha_{14} & 0 \\
	0  & 0 & \alpha_{13} & \alpha_{14} \\
    0  & \alpha_{23} & \alpha_{24} & 0 \\
    0  & 0 & \alpha_{23} & \alpha_{24}
\end{bsmallmatrix}}_\text{$\mathsf{C}_1$} u +
\underbrace{\begin{bsmallmatrix}
    \alpha_{12}  & \alpha_{15} & \alpha_{16} & 0 \\
	0  & \alpha_{12}  & \alpha_{15} & \alpha_{16} \\
    \alpha_{22}  & \alpha_{25} & \alpha_{26} & 0 \\
    0  & \alpha_{22}  & \alpha_{25} & \alpha_{26}
\end{bsmallmatrix}}_\text{$\mathsf{C}_0$}\Bigg)\boldsymbol{\mathsf{v}} = 0,$
}
\end{equation}
which is equivalent to the GEP:
\begin{equation}
\label{eq13}
\mathsf{A}\boldsymbol{\mathsf{y}}=u\mathsf{B}\boldsymbol{\mathsf{y}},
\end{equation}
with $\mathsf{A}=
\begin{bsmallmatrix}
0_4 & I_4\\ 
-\mathsf{C}_0 & -\mathsf{C}_1
\end{bsmallmatrix}$ and $\mathsf{B}=
\begin{bsmallmatrix}
I_4 & 0_4\\ 
0_4 & \mathsf{C}_2
\end{bsmallmatrix}$, $0_4$ and $I_4$ being the $[4\times 4]$ zero and identity matrices, and $\boldsymbol{\mathsf{y}}=
\begin{bsmallmatrix}
\boldsymbol{\mathsf{v}} \\
u\boldsymbol{\mathsf{v}}
\end{bsmallmatrix}=[\mathsf{y_1}\textmd{ }\mathsf{y_2}\textmd{ }\dots\textmd{ }\mathsf{y_8}]^T$. A particular solution $v_1$ is e.g., the quotient $\frac{\mathsf{y_3}}{\mathsf{y_4}}$ (or equivalently $\sqrt[3]{\frac{\mathsf{y_1}}{\mathsf{y_4}}}$) from the eigenvector associated with eigenvalue $u_1$. 

Note that the resolution of (\ref{eq13}) is two orders of magnitude faster compared to (\ref{eq11}) using linear algebra packages.

Lastly, when the two ellipses $E_1$ and $E_2$ intersect in two points, two out of the four eigenvalues obtained for $u$ are complex. These correspond to the two internal bitangents: in that case, ellipses have only two external bitangents associated with the other two real eigenvalues. It is also worth noting that, when they exist, internal bitangents are associated with the extremal (real) eigenvalues.

\subsubsection{Selecting one bitangent}
\label{select_bitangent}
\begin{figure}
\centering
\includegraphics[width=0.47\textwidth]{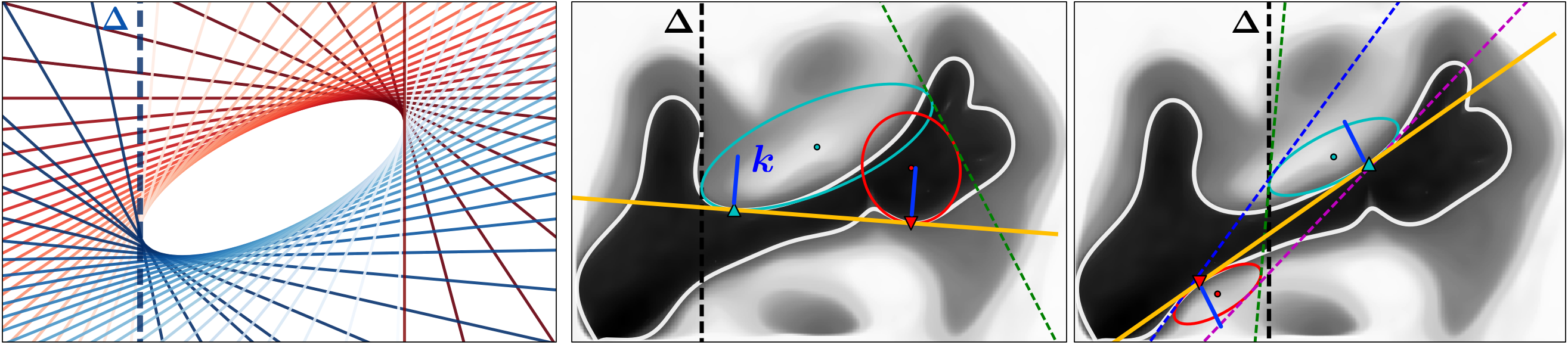}
\caption{Left: cyclic ordering of bitangents. Middle/right: Refinement of bitangents through ellipse fitting ($E_1$ is in cyan and $E_2$ in red). The curvature vectors are shown in blue, bitangent points are shown with triangles, and bitangents with coloured dashed lines. Selected refined bitangents are shown in yellow (usable bitangent types: middle, $LL$; right, $RL$).}
\label{fig:fig4}
\end{figure}

In this section, we identify the only bitangent of $E_1$ and $E_2$ that is also a bitangent of $\mathcal{L}$ (Fig.~\ref{fig:fig4})---referred to as the \textit{usable} bitangent.

Let us consider: $(i)$ bitangents directed from $E_1$ to $E_2$, $(ii)$ $E_1$ is oriented positively and $(iii)$ $\Delta$ is its left-most vertical tangent. Bitangents of $E_1$ can be cyclically ordered by considering independently the tangents below (in blue in Fig.~\ref{fig:fig4} Left), and above (in red) it, and sorting them by decreasing $y$-intercept with the ellipse's left-most tangent, $\Delta$. This holds for cases where an ellipse lies above (resp. below) all of the bitangents. Lemma 1 in \cite{habert2005computing} states that the resulting cyclic order of the bitangent directions is $\mathcal{C}$: $[LL,LR,RL,RR]$ ($L$ and $R$ stand for left and right and refer to the locations of an ellipse relative to a bitangent). 

Four possible cases arise: (c1) $E_2$ stands to the right of $E_1$, (c2) is above $E_1$ intersecting $\Delta$, (c3) is to the left of $E_1$, and (c4) is below $E_1$ intersecting $\Delta$. For each case, the first bitangent encountered starting from $\Delta$, counter-clockwise, has type $LL$, $RR$, $RL$ and $LR$ respectively; the next up to three bitangents for each case have their types deduced from the positive cyclic order $\mathcal{C}$.

Now in order to select the usable bitangent, one has to rely on the geometry of the level line $\mathcal{L}$. Let us define the unit curvature vector $\boldsymbol{k}$, at every point along $\mathcal{L}$ as the vector pointing toward the centre of the osculating circle: $\boldsymbol{k}=\kappa \boldsymbol{n}=\langle k_x, k_y \rangle$, where $\kappa$ is the scalar curvature and $\boldsymbol{n}$ is the normal (it is colinear to the gradient of the image along $\mathcal{L}$ and directed toward the inside of the clockwise-oriented closed curve here). The orientation of $\boldsymbol{k}$ allows differentiating otherwise ambiguous situations; for example, two pairs of ellipses ($E_1$, $E_2$) and ($E_1$, $E_3$), all of them fitting portions with same curvature and satisfying the configuration of case (c1), can be associated with a different type of usable bitangent, $RR$ and $RL$ respectively. This happens when $\boldsymbol{k_1}$ and $\boldsymbol{k_3}$ have opposite sense, while $\boldsymbol{k_1}$ and $\boldsymbol{k_2}$ have the same. In the following, positiveness is defined for (c1) and (c3) as $k_y>0$ and as $k_x>0$ for (c2) and (c4), and is denoted with the superscript $(+)$.


From there we define four patterns: (p1) $(k_1^{(+)},k_2^{(+)})$, (p2) $(k_1^{(+)},k_2^{(-)})$, (p3) $(k_1^{(-)},k_2^{(+)})$ and (p4) $(k_1^{(-)},k_2^{(-)})$. In cases (c1) and (c2), they correspond to the usable bitangent type $LL$, $LR$, $RL$, $RR$ respectively. Conversely, in cases (c3) and (c4), they correspond to the type $RR$, $RL$, $LR$, $LL$ respectively. Since there is a one to one correspondence between the four bitangents and the four types, it only requires identifying one of four patterns (p) and one of four cases (c) to pick the usable bitangent parameters.

We also extend the mapping to intersecting ellipses (Fig.~\ref{fig:fig4} Middle) by observing that the cyclic order of bitangents is of the form $[T_e, T_i, T_i, T_e]$ (subscripts $e$ and $i$ stand for external and internal). Since only external bitangents exist in the case where $E_1$ and $E_2$ intersect in two points (\text{\S}\ref{bitangent_ellipses}), we are left with the cyclic order $[T_e, \_, \_, T_e]$.

Bitangent points are straightforward to obtain for $E_1$ and $E_2$ by substitution of the tangent equation in the ellipses equations. Finally, we select the point of $\mathcal{L}$ that is the closest to an ellipse bitangent point. Note that once all bitangents are refined, some bitangent points may collapse to similar locations. In order to reduce ineffective redundancy, only one bitangent out of those that have their end points close to each other is kept \cite{rothwell1995object}.

\section{Projective shape representation}
\label{projective_representation}
We now have a set of refined bitangents. Let us consider one bitangent and its endpoints $b_1$ and $b_2$. In order to encode the shape of a portion of (oriented) level line $\mathcal{L}_r = \mathcal{L}\restriction[b_1,b_2]$ (assuming $b_1$ comes before $b_2$) in a projective invariant manner (as opposed to affine invariant \cite{lamdan1988object}, used in \cite{lisani2003theory}), two more points are required: the cast points $c_{.}$. The four points $b_1, c_1, c_2, b_2$, invariant under projective transformation, form the vertices of a polygon---the level line frame $\mathcal{F}_l$---and are mapped to the unit square vertices, $\mathcal{F}_c$ (the canonical frame) \cite{rothwell1995object}. The resulting projection is applied to $\mathcal{L}_r$ and provides a \textit{canonical} curve that can be used for shape comparison and matching.

A cast point $c_1$ (resp. $c_2$) is defined as the contact point of the tangent to $\mathcal{L}_r$ that intersects the level line at $b_1$ (resp. $b_2$). There exist several such points for each bitangent point in the case of long bitangents. It thus becomes critical to ensure that a candidate frame $\mathcal{F}_l$ forms a convex polygon so as to get an acceptable projection of $\mathcal{L}_r$ to the canonical frame. In the case of short bitangents, the construction of $\mathcal{F}_l$ is straightforward as only two cast points exist. As for long bitangents, a single portion of curve may be associated with several canonical curves, each of which depends on the frame configuration. As noted in \cite{rothwell1995object}, it is preferable to pick those making a wide angle between the bitangent and the cast tangents, as well as those having the cast points as far from one another as possible: unbalanced frames may give distorted canonical curves. This holds for bitangents crossing the level line. It is also worth mentioning that this step drastically prunes the set of bitangents that can lead to satisfying frames.

\begin{figure}
\centering
\includegraphics[width=0.47\textwidth]{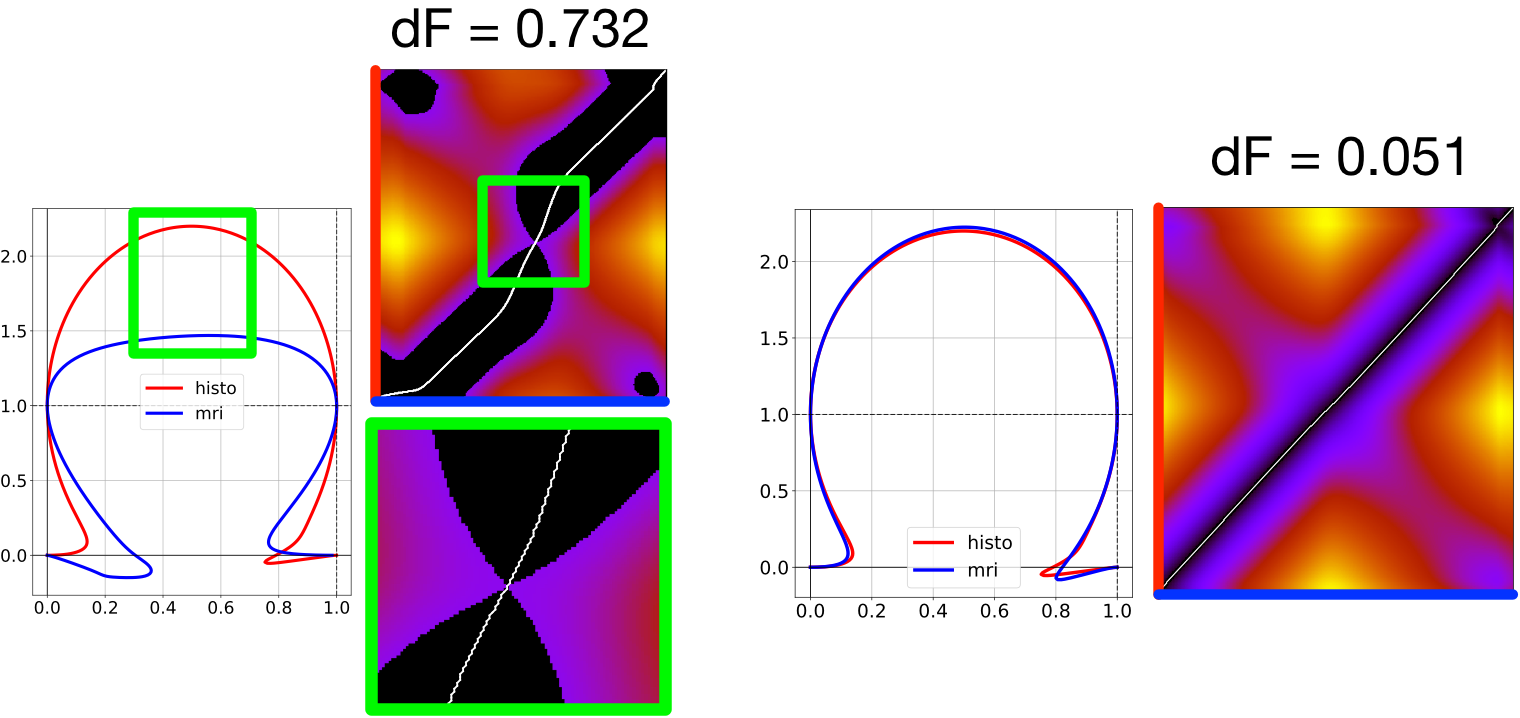}
\caption{Comparison of canonical curves (CC) and free space diagrams. For two shape elements, the Fr\'echet distance, $\textmd{dF}$ is computed between 2 CC from histology (red) and MRI (blue) and the associated free space diagrams with Fr\'echet paths (white line) are shown---only endpoints of segments are used. The regions in black correspond to the reachable free space ($\delta\leq \textmd{dF}$ here).}
\label{fig:fig5}
\end{figure}

\subsection{Canonical curves}
\label{cano_curves}

The goal is here to determine the 2D homography matrix such that $\mathbf{x_i}=\rho\mathsf{T}\mathbf{X_i}$ \cite{hartley2003multiple}, where $\mathbf{X_i} = [X_i\textmd{ }Y_i\textmd{ }1]^T$ is the~$i$-th point in $F_l$ (which no 3 are colinear) in homogeneous coordinates, $\mathbf{x_i} = [x_i\textmd{ }y_i\textmd{ }1]^T$ is the $i$-th vertex of the unit square defined by $(0,0,1)$, $(0,1,1)$, $(1,1,1)$ and $(1,0,1)$, $\mathsf{T}$ is a $[3\times 3]$ matrix of the transformation parameters with $\mathsf{T_{33}}=1$ and $\rho$ is a non-zero scalar that gives by elimination 8 equations from four correspondences, linear in the parameters. The solution we are seeking is the unit singular vector corresponding to the smallest singular value of the matrix of the system.

A normalisation step, which consists of translating and scaling, is recommended for it forces the entries of the matrix of the system to have similar magnitude. Further details can be found in \cite{hartley2003multiple} (p.108).

\subsection{Comparing polygonal curves}
\label{Frechet}
Contrary to \cite{rothwell1995object}, who relied on rays extended from an origin $(1/2,0)$ in $\mathcal{F}_c$ and designed a feature vector made of all the distances from every intersection point with the canonical curve to the origin, we compare canonical curves by means of the Fr\'echet distance (Fig.~\ref{fig:fig5}). The rationale is that we also consider bitangents that cross level lines. This means that the canonical curves may cross the base of $\mathcal{F}_c$ one or several times with more or less complex convolutions, making the use of rays impractical.

There are (at least) two common ways of defining the similarity between polygonal curves: the Hausdorff distance \cite{xia2004image} and the Fr\'echet distance. The latter has the advantage that it takes into account the ordering of the points along the curves, thereby capturing curves structure better \cite{alt1995computing}. For the sake of speed, we used the discrete Fr\'echet distance (see Table 1 of \cite{eiter1994computing}), which is an approximation of the continuous Fr\'echet distance: it only uses the curves vertices for measurements. From there, one can also define the reachable free space, which is the set of points for which the distance between two curves is lower than a distance parameter, $\delta$ and this allows tracking local similarity \cite{buchin2009exact}. The Fr\'echet distance is the minimum $\delta$ that allows reaching the top right corner of the free space starting from $(0,0)$.

\section{Matching and registration}
\label{reg}
By cross-comparisons between histology and MRI, one obtains a measure of shape similarity (\text{\S}\ref{Frechet}). Because each level line is associated with many canonical curves (one for each shape element), matches are found when the Fr\'echet distance is minimum and below a certain threshold. We can then use correspondences between $\mathcal{F}_l$ in histology and MRI to compute an affine transformation (same principle as in~\text{\S}\ref{cano_curves} with only 3 points---each providing two equations---and ${\boldsymbol{\mathsf{p}}=[T_{11}\textmd{ }T_{12}\textmd{ }\dots\textmd{ }T_{23}\textmd{ }0\textmd{ }0\textmd{ }1]^T}$). In order to minimise the global alignment error, the points must be well-arranged in images, i.e. the frames should be as wide as possible (hence the advantage of using long bitangents). When considering several level lines in both modalities, each canonical curve of each level line from histology returns at most one matching canonical curve for each level line in the MRI. False matches are filtered out using random sample consensus (RANSAC) \cite{fischler1981random} and a single global transformation is computed.

\section{Results and discussion}
\label{results}

We evaluated the method on 7 pieces of tissue, altogether covering 3 different subjects. For each subject, we had access to T2w, PSIR and PD MRI volumes (7 slices, 0.25$\times$0.25$\times$2mm$^3$). From these volumes we selected the slice that visually looked the most similar to one piece of histology. Histological images were a series of 11 consecutive 2$\mu m$-thick sections, stained with 11 different dyes. At this point, it is worth noting that because the histological slab was about 25$\mu$m-thick---compared to a 2mm-thick slice from MRI---projective invariance not only allowed being robust to tissue distortions in the recognition process but was also required in order to tolerate morphological variations happening within that 2mm gap.

A ground truth arrangement similar to that of Fig.~\ref{fig:fig1} was available for direct assessment of success or failure of the alignment. It was made by a histopathologist at the time of the tissue preparation and essentially consisted of reporting the cassettes locations onto a slice of a medical image in order to keep track on which part of the sample the tissue piece was cut from. In the following, we call \textit{confusing} (as opposed to \textit{meaningful} \cite{desolneux2001edge} i.e., the tissue outer/inner boundaries) level lines, those not providing relevant information about the tissue shape.

\begin{figure}
\centering
\includegraphics[width=0.37\textwidth]{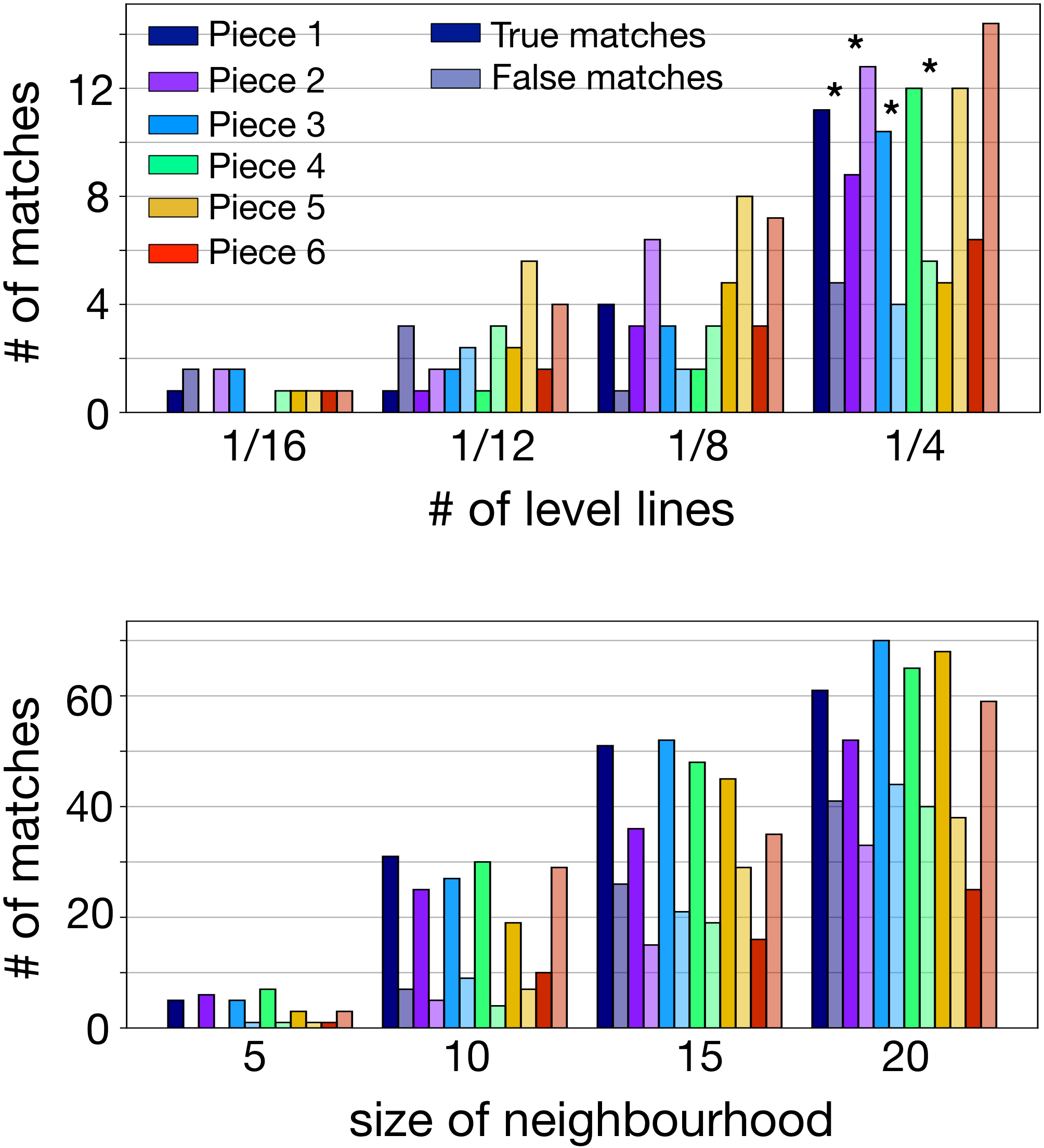}
\caption{Top: (E1) joint effects of sparsity and confusedness on the recognition of shape elements between histology and MRI for 6 pieces of tissue, along with the ability of RANSAC to provide the correct transformation (*). Bottom: (E2) effects of redundancy/confusedness. Numbers of true/false matches (different opacity) are reported for each piece (different colours) in both experiments. RANSAC is successful 5 out of 6 times in (E2).}
\label{fig:fig6}
\end{figure}

We ran two experiments (Fig.~\ref{fig:fig6}): (E1) consisted of using levels multiple of 16, 12, 8 and 4 in histology and MR images to investigate two questions: what is the impact of confusing level lines as well as their number, on the matching and the alignment? Can level lines be used as they are, without any form of prior knowledge about the tissue boundaries in images? Note that level lines were computed at quantised levels 0 to 255 by steps of 1. We expect that the sparser the set of level lines, the less informative about the actual tissue shape they can be (since information is lost when quantisation is coarse). This indeed translates in higher numbers of false than true matches when using between 1/16th and 1/8th of all available level lines (except for pieces~1 and 3 when using 1/8th, but this is hardly representative). When sufficient information comes in (1/4th), recognition becomes more successful: despite finding more false than true matches for piece~2, RANSAC was able to return the correct transformation---most of the false matches being isolated and spread across the MR image domain in that case. In contrary, RANSAC was unable to deal with false matches for pieces 5 and 6, those being related to ambiguities (shape elements were small and confusing).

The second experiment (E2) investigated the question: how robust is the matching/alignment when injecting confusing information into a subset of meaningful level lines? As such, we increased the number of neighbouring level lines from $\pm$5 to $\pm$20 around a meaningful one. In practice, meaningful level lines are those around structural layers (contrasted boundaries) of the tissue and we manually picked the corresponding levels. We can observe that the more localised around relevant information the level lines are, the higher the ratio true/false matches and the more trustful the set of correspondences fed to RANSAC. This is where redundancy is very valuable. However, the more levels one includes, the further one goes from meaningful information, and the more confusing it can get (see the increase in false matches). Due to the complexity of the information and the sinuosity of the shape, we believe that starting from a meaningful subset of level lines is an important consideration.

\begin{figure}
\centering
\includegraphics[width=0.45\textwidth]{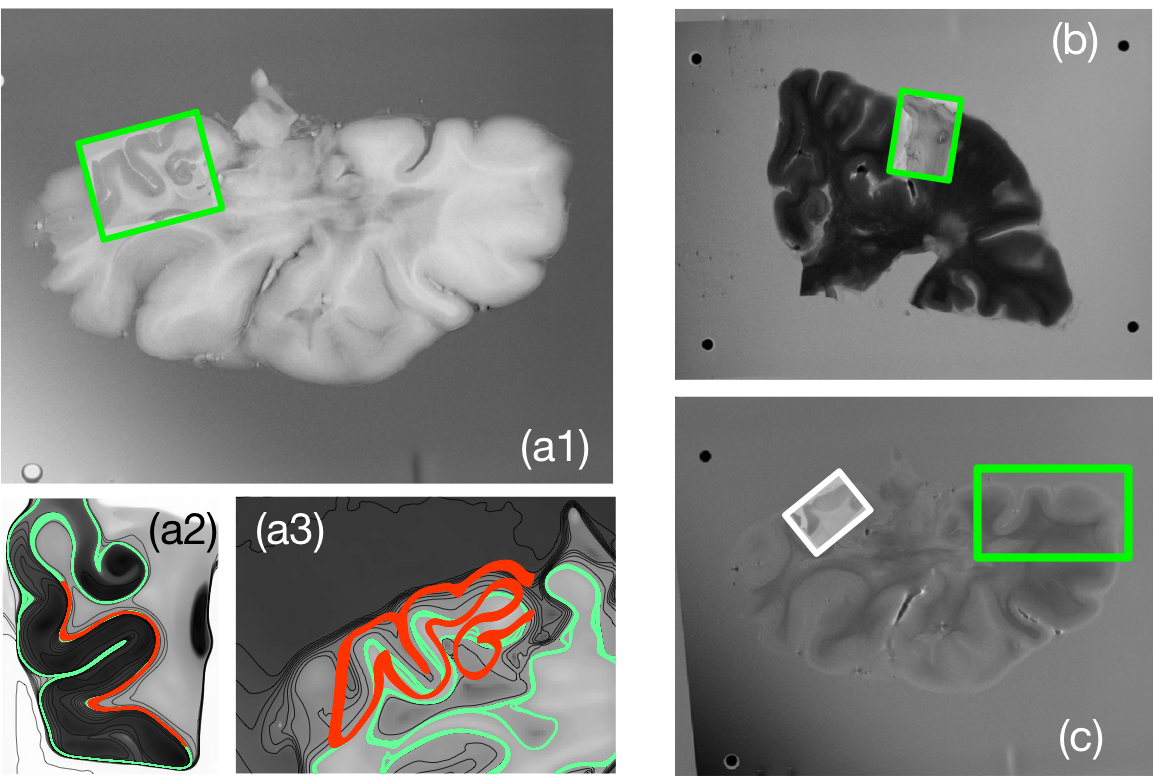}
\caption{Alignment results. (a1)-(b)-(c) Successes and failure of the approach for pieces~1, 5 and 6, using PSIR, T2 and PD images respectively. (a2) Example of matching shape elements (orange) and associated level lines (green and black) of piece~1. (a3) Affine-transformed matching level lines of histology (orange) overlaid onto matching level lines (green) of PSIR and its other level lines (black).}
\label{fig:fig7}
\end{figure}

Resulting alignments are shown in Fig.~\ref{fig:fig7} for 3 pieces, considering neighbourhoods of $\pm$10 level lines. Overall, 5 pieces were matched correctly and two incorrectly. As for piece~6, no shape element was discriminative enough to be correctly matched with an MRI portion of level line without any ambiguity (Fig.~\ref{fig:fig7}c), as only relatively short bitangents could be extracted. As for piece~7, this is due to the fact that it is close to convex (and thus was not considered in the previous experiments). As a result, a few or no bitangents could be extracted from that histological image and no match was therefore available. 

The main requirements of the approach are twofold and relate to the \textit{length} of the bitangents and the threshold on the Fr\'echet distance. As stressed out earlier, short bitangents convey little and ambiguous information about shape. This results in false matches especially because of the tolerance of the projective-invariant setting and the sinuosity of the MRI level lines. As a matter of fact, we constrained the approach to using long bitangents: in practice, we used those covering portions of a level line with more than 6 inflections. If a histological image happened to have informative portions with more than two inflections but less than 6---as it was the case for piece 5---then the longest bitangents were used (4 and 5 inflections in that case). An upper bound was also set (we chose 10 inflections) in order to speed up the matching process and avoid aberrant comparisons with bitangents covering the whole MR image; that range was applied to both MRI and histology. The rationale for considering such a range is also that it is not guaranteed that two level lines have the exact same number of inflections on corresponding portions across modalities, but their smoothness ensures those numbers are close. Long bitangents produce characteristic canonical curves (furthermore associated with wide frames) and allow for lower thresholds on the Fr\'echet distance while discarding false matches better.

\section{Conclusion}
\label{discussion}
This paper stands as a proof of concept that multimodal registration between a piece of tissue from histology and its whole in an MRI---which, to the best of our knowledge, remains to be addressed---is achievable as a result of shape recognition using portions of their level lines. Such a formulation allows for contrast, projective invariant representation of shape elements and partial matching regardless of the orientation of the piece of tissue on the glass slide (flips, rotations). We also introduced a computationally efficient refinement of bitangents using ellipses, from which a single bitangent was retained according to the local geometry of the level line. All this however, is to be related with the complexity of medical images; successful alignments require subsets of meaningful level lines along with characteristic shape elements. Those were obtained via the extension of Rothwell's framework to bitangents crossing the level lines and by preferring long bitangents.

Future works include: $(i)$ the automatic extraction of meaningful level lines \cite{desolneux2001edge}; $(ii)$ the use of shortcut Fr\'echet distance \cite{driemel2013jaywalking}, which bypasses large dissimilarities. This could improve robustness to tissue tears: a level line in histology may be globally close in terms of its shape to part of another in MRI but because of a tear that it follows, the distance between the associated canonical curves will be large.

\section*{Acknowledgments}
The authors would like to thank Prof. Olga Ciccarelli and her group (UCL Institute of Neurology, Queen Square MS Centre), for kindly providing the data.


This research was supported by the European Research Council (Starting Grant 677697, project BUNGEE-TOOLS), the University College London Leonard Wolfson Experimental Neurology Centre (PR/ylr/18575), the Alzheimer's Society UK (AS-PG-15-025), the EPSRC Centre for Doctoral Training in Medical Imaging (EP/L016478/1), the National Institute for Health Research University College London Hospitals Biomedical Research Centre and Wellcome/EPSRC (203145Z/16/Z, NS/A000050/1).

{\small
\bibliographystyle{ieee}
\bibliography{egbib}
}

\end{document}